\pdfoutput=1

\documentclass[11pt]{article}

\usepackage{acl}

\usepackage{times}
\usepackage{latexsym}

\usepackage[T1]{fontenc}

\usepackage[utf8]{inputenc}

\usepackage{microtype}

\usepackage{inconsolata}

\usepackage{graphicx} 
\usepackage{algorithm}
\usepackage{algpseudocode}
\usepackage{amsmath}
\usepackage{amsfonts}
\usepackage{times}
\usepackage{latexsym}
\usepackage{bm}
\usepackage{booktabs} 
\usepackage{anyfontsize}
\usepackage[T1]{fontenc}
\usepackage{xspace}
\usepackage{enumitem}

\usepackage[utf8]{inputenc}

\usepackage{microtype}

\usepackage{inconsolata}

\newcommand{\method}{\texttt{TTM-RE}\xspace}

\title{\method: Memory-Augmented Document-Level Relation Extraction}

\author{Chufan Gao$^1$, Xuan Wang$^{2\dagger}$, Jimeng Sun$^{13\dagger}$\\ 
$^1$University of Illinois Urbana-Champaign $^2$Virginia Tech\\ 
$^3$Carle Illinois College of Medicine\\ 
\texttt{\{chufan2,jimeng\}@illinois.edu, xuanw@vt.edu}
}

\begin{document}
\maketitle
\def\thefootnote{$\dagger$}\footnotetext{Equal Senior Contribution}\def\thefootnote{\arabic{footnote}}

\begin{abstract}
Document-level relation extraction aims to categorize the association between any two entities within a document.
We find that previous methods for document-level relation extraction are ineffective in exploiting the full potential of large amounts of training data with varied noise levels. 
For example, in the ReDocRED benchmark dataset, state-of-the-art methods trained on the large-scale, lower-quality, distantly supervised training data generally do not perform better than those trained solely on the smaller, high-quality, human-annotated training data.
To unlock the full potential of large-scale noisy training data for document-level relation extraction, we propose \method, a novel approach that integrates a trainable memory module, known as the Token Turing Machine, with a noisy-robust loss function that accounts for the positive-unlabeled setting.
Extensive experiments on ReDocRED, a benchmark dataset for document-level relation extraction, reveal that \method achieves state-of-the-art performance (with an absolute F1 score improvement of over 3\%). Ablation studies further illustrate the superiority of \method in other domains (the ChemDisGene dataset in the biomedical domain) and under highly unlabeled settings.
\end{abstract}

\section{Introduction}
\begin{figure}[t]
    \centering
    \includegraphics[width=\linewidth]{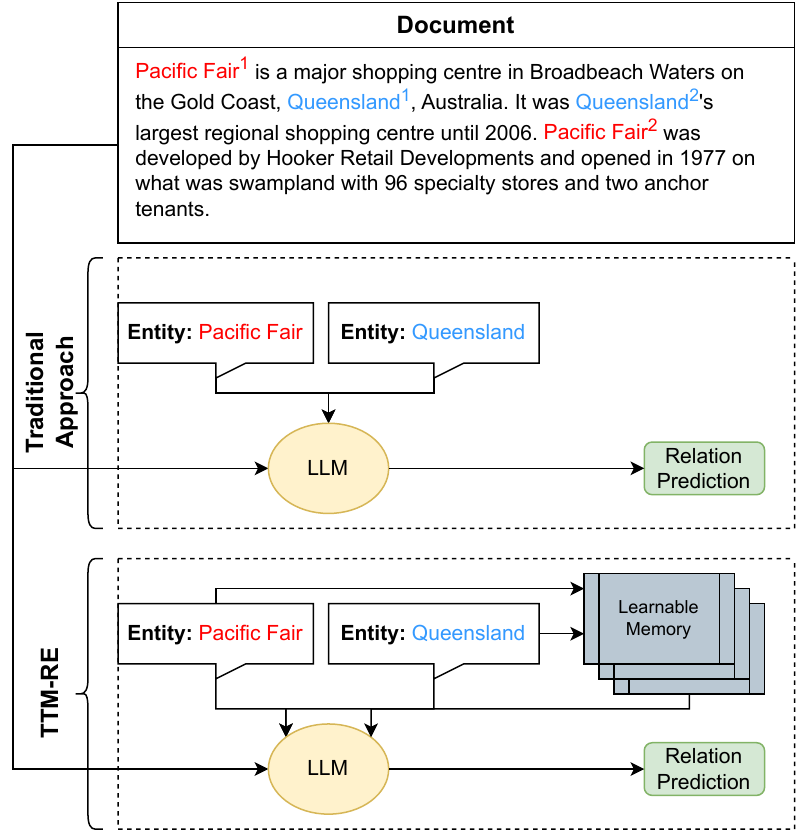}
    \caption{Differences between the generic document relation extraction approach and \method for document-level relation extraction. The memory module processes the input entities and outputs to the relation classifier. We investigate how adding the memory component affects performance (such as different datasets and memory sizes). 
    }
    \vspace{-1em}
    \label{fig:zerore_example}
\end{figure}

Relation extraction aims to classify the relationships between two specified entities into predefined categories. 
This task is pivotal in natural language processing, as it involves identifying and categorizing the connections between two entities (for example, "Pacific Fair" and "Queensland" in Figure \ref{fig:zerore_example}) into predefined classes (for example, "located in" in Figure \ref{fig:zerore_example}).
Its importance spans across various downstream applications, encompassing question answering \cite{veena2017graphQA}, knowledge graph construction \cite{RE_for_IR}, and the extraction of general patterns \cite{han2020more}.

Previous work for relation extraction mainly focuses on sentence-level relations \cite{alt2020tacred}.
For example, \citet{sainz2021label} characterized each relation class using a label verbalizer and addressed the relation extraction task through a textual entailment model, as well as other models such as DeepStruct \cite{wang2022deepstruct}. 
However, techniques that are primarily designed and evaluated for extracting relationships at the sentence level, face challenges that limit their suitability for datasets focused on document-level relation extraction such as ReDocRED \cite{tan2022revisiting}. 

These challenges include a large label imbalance and a large number of possible combinations between possible head and tail entity pairs in each document, which is quadratic.
Previous work has generally addressed DocRE's label imbalance with custom loss functions and its quadratic entity computation by minimizing document processing.
Usually, a common processing step for the document inspired by \citet{zhou2021documentadaptiveThre} is used. 
Initially, a pre-trained encoder processes the entire document as a whole. Subsequently, indexing over the entity mentions can be employed to retrieve the head and tail entities for classification.
Recent work has mainly focused on novel loss functions \cite{, tan2022documentAdaptiveFocalLoss, wang2022unified} or additional inputs such as evidence \cite{ma2023dreeam}. 

However, less effort has been made in effectively leveraging the large amount of the distantly labeled data in ReDocRED and DocRED \cite{yao2019docred, tan2022revisiting}. 
Most work \cite{tan2022document, ma2023dreeam} use the distantly labeled dataset for knowledge distillation--that is, a model is first trained on the fully human annotated train data, and then used to obtain output logits on the distantly supervised data. These logits, along with the distant relation labels, are then used as guidance in training a secondary student model. 
However, previous work investigating fine-tuning on the distantly supervised dataset has failed to significantly boost performance (as we show later in our results). Previously, this could be explained by the lower quality and the lack of human annotation on the distantly supervised dataset. However, we assert that this is due to an architectural limitation on the prevailing framework.

Recently, numerous studies have highlighted the advantages of incorporating memory in both computer vision and NLP to acquire pertinent representations from past data points that facilitate improved classification performance.
For instance, \citet{barraco2023little} showcased the enhanced performance of an encoder-decoder model for image captioning by integrating memory of past observations into the attention mechanism. 
Additionally, the integration of memory has been demonstrated to enhance performance in knowledge-intensive tasks such as long-form question answering and dialogue \cite{wu2022efficient}. 
More recently, Google's Token Turing Machine (TTM) \cite{ryoo2023token} has showcased state-of-the-art performance in real-world long-sequential visual understanding using an autoregressive Transformer model equipped with memory.

Drawing from recent advancements in memory-augmented models, we introduce \method, the inaugural memory-augmented architecture designed specifically for document-level relation extraction. 
Through empirical evidence, we demonstrate that this architecture enables notably enhanced fine-tuning on extensive distantly labeled data from empirical observations.
Specifically, we show that adding memory tokens from TTM \cite{ryoo2023token} empirically enhances downstream relation classification by allowing reprocessing of head and tail entities while jointly considering the learned memory tokens in mind (see Figure~\ref{fig:zerore_example}).

\begin{enumerate}[leftmargin=*]
    \item We propose \method, the first memory-augmented document-level relation extraction model. By incorporating pseudo entities, it significantly enhances downstream relation classification performance on datasets such as ReDocRED (+3 F1 score) and ChemDisGene (+5 F1 score).
    \item We show that without any human-labeled data, \method achieves impressive relation extraction performance on unseen data (+9 F1 score). Furthermore, in an extremely unsupervised scenario (19\% of training labels), \method outperforms the previous SOTA by an impressive margin (+12 F1 score).  
    \item We perform a thorough analysis of ablations examining the performance of \method across less/more frequent relation classes, assessing the impact of memory size, layer size, and the utilization of different base models.
\end{enumerate}

\begin{figure}[t]
  \texttt{{\color{red}Pacific Fair$^1$} is a major shopping centre in Broadbeach Water on the Gold Coast, {\color{blue}Queensland$^1$}, Australia. It was {\color{blue}Queensland$^2$}'s largest regional shopping centre until 2006. {\color{red}Pacific Fair$^2$} was developed by Hooker Retail Developments and opened in 1977 on what was swampland with 96 specialty stores and two anchor tenants.}
  \caption{Sample document relation extraction document from DocRED \cite{yao2019docred}. Here, the \textcolor{red}{head entity} is related to the \textcolor{blue}{tail entity} by \texttt{"P131: located in the administrative territorial entity"}. 
  }
  \label{fig:sample}
\end{figure}
\section{Related Work}


\subsection{Document-level Relation Extraction}
Document-level relation extraction stands as a critical endeavor in natural language processing, given that over 40.7\% of relations necessitate the extraction of information spanning multiple sentences and multiple entity mentions \cite{yao2019docred, tan2022revisiting, zhang-etal:2022:LREC}. In Figure~\ref{fig:sample}, we illustrate an instance of document-level relation extraction. Here, the objective is to discern the relationship between a pair of entities ("Pacific Fair" and "Queensland") within the provided document. Each entity is referenced twice in the text (indicated by superscripts).

Thus, a document with $n$ entities will have $n(n-1)$ possible relation predictions. \citet{zhou2021documentadaptiveThre} proposed ATLOP, which uses a single pass to encode the document using Roberta-large and indexing relevant entities in the token embeddings. Many works \cite{tan2022documentAdaptiveFocalLoss, wang2022unified, tan2022document} have proposed new loss functions on top of a single encoder pass (usually added on top of ATLOP \cite{zhou2021documentadaptiveThre}) to alleviate the combinatorial bottleneck and propose loss functions to tackle the class-imbalance problem in document-level relation extraction.  \citet{zhou2021documentadaptiveThre} introduced the ATLOP, which uses Roberta-large as the encoder and adaptive thresholding for the multilabel relation classification. \citet{tan2022documentAdaptiveFocalLoss} introduced KD-DocRE, which combines axial attention over the entity mentions with adaptive focal loss and knowledge distillation. \citet{wang2022unified} introduced a loss that considered a positive-unlabeled nature of the label distribution. \citet{ma2023dreeam} proposed using the evidence labels along with the distantly supervised labels to achieve the current SOTA. Like other works, \method also utilizes this single-pass encoder approach, as it is elegant and efficient. However, we introduce a new model component described in the next section.

\subsection{Memory-based Models in NLP}
Memory-based models have begun to see rising usage in the CV and image captioning areas. However, their usage in NLP has been surprisingly limited. Still, there are some interesting and relevant work to our application.
\citet{de2021mention} utilized 'mention memory' to represent knowledge—a table of dense vector representations of each entity mention in the corpus, and demonstrated good performance on multiple open-domain tasks including claim verification, and entity-based QA.
\citet{zhong2022training} introduced a training objective that directly utilizes memory sets from local, long-term, and external and showed reduced perplexity on WIKITEXT-103.
\citet{chen2022augmenting} introduced a question-answer augmented encoder-decoder model and pretraining strategy, demonstrating improved performance on single-hop and multi-hop QA datasets.
\citet{wu2022efficient} learns keys and values that represent questions and corresponding answers respectively; at inference time, the model would retrieve information from the memory using maximum inner product search.

Inspired by Token Turing Machines, \method's memory mechanism differs from these approaches in that it does not necessitate learning relevant portions of real text. It simply learns memory tokens (dense embeddings). This is uniquely suited to our application, as retrieving and reprocessing text requires additional LLM encoder calls for each entity-entity pair, which is quadratic in nature.


\section{Methodology}
\label{sec:Methodology}

\begin{figure*}[ht]
    \centering
    \includegraphics[width=\linewidth]{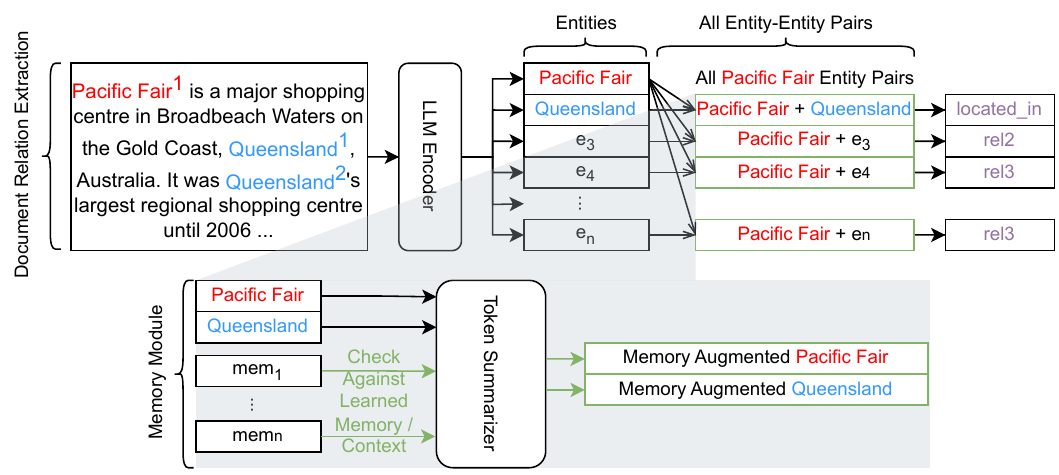}
    \caption{Overall framework of \method. Given an example document and an expected relation distribution, we use an LLM (Roberta-Large) to encode the input tokens in a single pass and consider \textcolor{red}{head} and \textcolor{blue}{tail} entities by their token representations, which are then fed into a memory module (\textcolor{gray}{in gray}). The memory module then returns 2 memory-augmented versions of the \textcolor{red}{head} and \textcolor{blue}{tail} entities for final relation classification.
    }
    \label{fig:zerore}
\end{figure*}

We propose \method, a memory-augmented, document-level relation extraction method. An illustration of the overall framework of \method is shown in Figure~\ref{fig:zerore}.


\subsection{Problem Definition}
In the task formulation, we examine a document $D$ that comprises $M$ sentences ($s_1$, $s_2$, ..., $s_M$), $N$ entities ($e_1$, $e_2$, ..., $e_N$), and $R$ relation classes. Given this document $D$ and a specified pair of entities ($e_{h}$, $e_{t}$), 
the goal is to forecast a set of positive relations ($\hat{r}_1, \hat{r}_2, ..., \hat{r}_p $) between the entity pair based on the information derived from the document. It's important to note that each entity may appear multiple times within the document $D$ and that each possible entity-entity pair needs to be considered (i.e. if $n$ entities, we need to consider $R \times N \times (N-1)$ possible relations).

\subsection{Token Turing Machines}
The memory, denoted as $M \in \mathbb{R}^{m \times d}$, comprises a collection of $m$ tokens each with a dimensionality of $d$. The input, consisting of $n$ tokens represented by $I \in \mathbb{R}^{n \times d}$, is combined with the memory $M$. This concatenated input is then further processed to generate an output denoted as $O \in \mathbb{R}^{r \times d}$, where $r$ represents the desired number of retrieved tokens. The outputs from this process, in conjunction with the preceding inputs and the current memory, constitute the output of the TTM. In our case, $I \in \mathbb{R}^{2 \times d}, O \in \mathbb{R}^{2 \times d}$ for the head and tail entities.

Token Turing Machines add support for external memory in the form of tokens (Figure~\ref{fig:zerore} Memory Module).
In Token Turing Machines (TTMs), the interface between the processing unit and memory are done purely in terms of ``read'' and ``write'' operations.
Note that in the original paper, the output from the processing unit is ``written'' to the memory, but in our case, since we are not applying this model sequentially, we can ignore this step and focus solely on the reading portion.

\paragraph{Initializing Memory Tokens:} We follow the original implementation of TTM and initialize memory tokens from scratch, with one major difference. While the original code initialized tokens from zeros, we found that this led to a lack of gradient updates. Therefore, we initialize from a normal distribution to allow for improved learning. Note that we cannot simply use entity text embeddings since the memory layer is after all processing steps and only before the final classification layer. We found this setup to work the best empirically, but further research is needed on the placement of the memory mechanism.

\paragraph{Reading from Memory:}
While the memory is intended to encapsulate condensed information deemed significant by the model, not all of this data may be relevant. Additionally, redundancies in the input, denoted as $I$, can arise due to information already stored in our memory, $M$, or inherent within the data itself. Selective reading where only a smaller subset of tokens is considered should encourage the model to create a memory repository containing relevant information over the entire relation classification task. 

We summarize a token set $\bm{I} \in \mathbb{R}^{n \times d}$ by deriving an importance weight vector, $w_i \in \mathbb{R}^{n}$, which we utilize to compute a weighted aggregation across the $n$ tokens. Notably, each output token, indexed as $i \in {1, \ldots, k }$, possesses its corresponding weight $w_i$, computed using a learnable function that takes the input $\bm{I}$ itself, denoted as $\alpha_i(\bm{I}) : \mathbb{R}^{d} \rightarrow \mathbb{R}$.
This importance weighting function is implemented through a Multi-Layer Perceptron (MLP) determined as $w_i = \alpha_i(\bm{I}) = \text{softmax}(\text{MLP}(\bm{I}))$.

Subsequently, these weights facilitate a weighted summation of the inputs. Let $\bm{V}$ be a generic list of input tokens we wish to analyze. For our specific case, it is the concatenation of both the memory tokens $\bm{M}$ and the input tokens $\bm{I}$. I.e. $\bm{V} = [\bm{M}\| \bm{I}]$. Let us obtain encoded token
$\bm{z}_i = w_i \cdot \bm{V} = \alpha_i(\bm{V}) \cdot V$,
where each token $\bm{z}_i$ effectively condenses all tokens from the complete set $\bm{V}$, guided by the dynamic weighting $w_i = \alpha_i(\bm{V})$.
As the model learns to summarize $p$ tokens into $r$ tokens, it generates a matrix $\bm{W} = [w_1, \cdots, w_r]$ comprising importance weights relative to the memory tokens. Also, to allow the model to read from a location, take advantage of memory token position, and distinguish them from input tokens, there is a learnable positional embedding \cite{dosovitskiy2020} before each read module. 
All of this is captured in a memory reading function defined as:
    $Z = \text{Read}(\bm{M}, \bm{I}) = S_r([\bm{M} || \bm{I}])$.
Where $[\bm{M} || \bm{I}]$ is the concatenated memory $\bm{M}$ and input $\bm{I}$, and $S_r(\cdot) : \mathbb{R}^{(|\bm{I}|+|\bm{M}|) \times d} \to \mathbb{R}^{r \times d}$. 
In our application, we set $r=2$ to learn memory augmented head $e_{h'} \in \mathbb{R}^d$ and tail $e_{t'} \in \mathbb{R}^d$ entities for the head and tail entity relation classification problem.

\subsection{Processing of Head and Tail Entities}
\label{sec:method_process}
After retrieving our memory-augmented head $S_{1}$ and tail $S_{2}$ entities, we 
use the group bilinear approach as specified in \citet{zhou2021document, tang2019orthogonal} to reduce the number of parameters to enable more efficient learning. Each entity is split into $k$ sections of dimension $d/k$, $e_{h'} = [e_{h}^1 || \dots || e_{h'}^k], e_{t'} = [e_{t'}^1 || \dots || e_{t'}^k]$
\begin{equation}
\begin{aligned}
\label{eq:group_bilinear}
    p(r | e_{h'}, e_{t'}) &s= \sigma\left( \sum_{i=1}^k e_{h'}^i B^i e_{t'}^i\right) \nonumber\\
\end{aligned}    
\end{equation}
$B^i \in \mathbb{R}^{d/k \times d/k}$ denotes bilinear layers, and the sum of the products represents the grouped bilinear layer. This reduces parameters from $d^2 \rightarrow d^2/k$ and enables much better performance empirically. 

Furthermore, the final output is a prediction vector of dimension $R+1$ (number of all relations + 1 to learn the threshold value), as we adopt the adaptive thresholding approach implemented by \citet{zhou2021document}, which most other recent Document RE work as done as well \cite{ma2023dreeam, tan2022documentAdaptiveFocalLoss}.

\subsection{Noise-Robust Loss Function (SSR-PU)}
There exists a large number of false negatives in the labeled relation triples. \citet{gao2023promptre} demonstrated difficulty in learning to ignore the false negatives for zero-shot prompting, revealing the difficulty of prompting LLMs for document relation extraction. To address this problem, we adopt Positive Unlabeled (PU) learning with the prior shift for each class as in \citet{wang2022unified} \cite{pmlr-v37-plessis15, NIPS2014_35051070}.

Ordinary PU learning assumes that the overall distribution is the same as the distribution of the unlabeled data, which may not be true in our case. To address this problem, PU learning under the prior shift of training data needs to be considered \cite{charoenphakdee2019positive}.
For each class, assume that the original prior $\pi_{i}=p(y_{i}=+1)$. Let $\pi_{labeled,i}=p(s_{i}=+1)$ and $(1-\pi_{labeled,i})=(1-p(s_{i}=+1))=p(s_{i}=-1)$ where $s_{i}=+1$ or $s_{i}=-1$ mean that the $i$-th class is labeled or unlabeled, respectively. 

The conditional probability of a positive sample under unlabeled data is:
\begin{equation}
\begin{aligned}\label{eq9}
\pi_{u,i} &= p(y_{i}=1 \mid s_{i}=-1) \\ \nonumber
&= \frac{\pi_{i}-\pi_{labeled,i}}{1-\pi_{labeled,i}}
\end{aligned}
\end{equation}

The non-negative risk estimator under class prior shift of training data is obtained as follows \citep{NIPS2017_7cce53cf, wang2022unified}:
\begin{equation}
\begin{aligned}\label{eq13}
&\widehat{R}_{\mathrm{S-PU}}(f)=\sum_{i=1}^{K}( \frac{\pi_{i}}{n_{\mathrm{P}_{i}}} \sum_{j=1}^{n_{\mathrm{P}_{i}}}\ell(f_{i}(\boldsymbol{x}_{j}^{\mathrm{P}_{i}}), +1) \\ \nonumber
&+\mathrm{max}(0, [\frac{1}{n_{\mathrm{U}_{i}}} \frac{1-\pi_{i}}{1-\pi_{u,i}} \sum_{j=1}^{n_{\mathrm{U}_{i}}} \ell(f_{i}(\boldsymbol{x}_{j}^{\mathrm{U}_{i}}), -1)\\&-\frac{1}{n_{\mathrm{P}_{i}}}\frac{\pi_{u,i}-\pi_{u,i} \pi_{i}}{1-\pi_{u,i}} \sum_{j=1}^{n_{\mathrm{P}_{i}}}\ell(f_{i}(\boldsymbol{x}_{j}^{\mathrm{P}_{i}}), -1)]))
\end{aligned}
\end{equation}
where $\pi_{i} = p(y_i = +1)$ denotes probability of positive prior for relation class $i$. $n_{P_i}$ are the number of positive and $n_{U_i}$ are the unlabelled samples of
class i, respectively. $\ell$ is a convex loss function, and $f_i(\cdot)$ is a score function that predicts class $i$. $\boldsymbol{x}_{j}^{\mathrm{P}_{i}}$ and $\boldsymbol{x}_{j}^{\mathrm{U}_{i}}$ denotes that the $j$-th sample of class $i$ is positive and unlabeled as class $i$ respectively. Please see Appendix~\ref{app:ssrpu_loss} for more details.

\section{Experimental Settings}
\subsection{Datasets}
\paragraph{ReDocRED}
\begin{table}[t]
\centering
\caption{Statistics of the Re-DocRED dataset (Train, Dev, and Test are fully reprocessed from DocRED for improved accuracy and completeness). In total, there are 96 relations. The distantly supervised dataset is the same as in DocRED and is created with no human supervision.}
\label{tab:data_stats}
\resizebox{\linewidth}{!}{%
\begin{tabular}{ccccc}
\toprule
Statistics                  & Distant & Train & Dev  & Test \\ \hline
\# Docs           & 101,873 & 3,053 & 500  & 500  \\
Avg. \# Entities  & 19.29 & 19.4  & 19.4 & 19.6      \\
Avg. \# Labeled Triples & 14.79 & 28.1  & 34.6 & 34.9      \\
Avg. \# Sentences & 8.13 & 7.9   & 8.2  & 7.9        \\ \bottomrule
\end{tabular}
}
\end{table}

To evaluate our methodology, we primarily use ReDocRED \cite{tan2022revisiting}, an open-access, document-level relation extraction dataset that improves upon the popular DocRED dataset \cite{yao2019docred} by resolving incompleteness, addressing logical inconsistencies, and correcting coreferential errors. 
Table~\ref{tab:data_stats} shows the amount of training data available for all data splits as well as the average number of entities. Note that we primarily use the Dev set of ReDocRED for our experiments for computational practicality.

\paragraph{ChemDisGene}
\begin{table}[t]
\centering
\caption{Statistics of the ChemDisGene dataset. In total, there are 14 relations. The distantly supervised training is created with no human supervision.}
\label{tab:chemdisgene_stats}
\resizebox{.8\linewidth}{!}{%
\begin{tabular}{ccc} \toprule
Statistics & Train (Distant) & Test \\ \midrule
\# Docs     & 76942           & 523  \\
Avg \# Entities & 7.5             & 10   \\
Avg \# Labeled Triples & 2.1             & 7.2  \\ \bottomrule
\end{tabular}
}
\vspace{-1em}
\end{table}
ChemDisGene \cite{zhang-etal:2022:LREC} is a biomedical multi-label document RE dataset. Entity mentions were obtained using PubTator Central \cite{wei2019pubtator}, and the relationships are based on the Comparative Toxicogenomics Database \cite{davis2021comparative}. Table~\ref{tab:chemdisgene_stats} shows the stats for the data.
It comprises 523 abstracts meticulously curated by a team of biologists. Our training uses the larger distantly supervised training set, while evaluation is conducted using the fully expert-labeled test set. The average number of relations per document in the test set across both datasets significantly exceeds the average number of relations per document in the training set.
This indicates the incomplete labeling phenomenon in the training set with a large number of false negatives, much like DocRED, before the updated ReDocRED.

\begin{table*}[t]
\centering
\caption{We compare all results against strong document relation extraction baselines: DREEAM \cite{ma2023dreeam}, ATLOP \cite{zhou2021document}, KD-DocRE \cite{tan2022documentAdaptiveFocalLoss}, and SSR-PU \cite{wang2022unified}. Bold denotes best performance or within 1 standard deviation.}
\label{tab:results}
\resizebox{\textwidth}{!}{%
\begin{tabular}{ccccccccc}
 & \multicolumn{4}{c}{Dev} & \multicolumn{4}{c}{Test} \\
Model & F1 & Ign F1 & Precision & Recall & F1 & Ign F1 & Precision & Recall \\ \toprule
\multicolumn{9}{c}{\textbf{Original (Human Annotation Only)}} \\ \midrule
DREEAM & 79.42$_{\pm0.18}$ & 78.36$_{\pm0.19}$ & 74.74$_{\pm0.64}$ & 85.15$_{\pm0.25}$ & 80.20$_{\pm0.45}$ & 78.56$_{\pm0.39}$ & 75.74$_{\pm0.65}$ & 83.89$_{\pm0.61}$ \\
ATLOP & 76.15$_{\pm0.23}$ & 75.88$_{\pm0.23}$ & 69.62$_{\pm0.81}$ & 84.26$_{\pm0.97}$ & 77.81$_{\pm0.71}$ & 76.13$_{\pm0.28}$ & 67.76$_{\pm0.23}$ & 85.35$_{\pm0.62}$ \\
KD-DocRE & 77.88$_{\pm0.42}$ & 77.12$_{\pm0.49}$ & 85.16$_{\pm0.58}$ & 71.30$_{\pm0.79}$ & 78.28$_{\pm0.72}$ & 77.60$_{\pm0.25}$ & 89.76$_{\pm0.14}$ & 69.40$_{\pm0.03}$ \\
SSR-PU & 78.58$_{\pm0.11}$ & 78.08$_{\pm0.14}$ & 75.59$_{\pm0.27}$ & 86.89$_{\pm0.51}$ & 80.18$_{\pm0.31}$ & 78.61$_{\pm0.46}$ & 69.43$_{\pm0.43}$ & 90.50$_{\pm0.53}$ \\
\method & 78.13$_{\pm0.12}$ & 78.05$_{\pm0.17}$ & 83.28$_{\pm0.29}$ & 76.28$_{\pm0.61}$ & 79.95$_{\pm0.13}$ & 78.20$_{\pm0.34}$ & 85.81$_{\pm0.55}$ & 76.68$_{\pm0.22}$ \\ \midrule
\multicolumn{9}{c}{\textbf{Distant Only}} \\ \midrule
ATLOP & 40.42$_{\pm0.61}$ & 32.14$_{\pm0.41}$ & 31.11$_{\pm0.66}$ & 60.30$_{\pm0.81}$ & 53.42$_{\pm0.73}$ & 51.14$_{\pm0.66}$ & 51.11$_{\pm0.69}$ & 55.95$_{\pm0.91}$ \\
SSR-PU & 39.35$_{\pm0.46}$ & 35.04$_{\pm0.38}$ & 23.35$_{\pm0.27}$ & 72.63$_{\pm0.47}$ & 54.46$_{\pm0.48}$ & 53.26$_{\pm0.20}$ & 48.02$_{\pm0.34}$ & 62.89$_{\pm0.42}$ \\
\method$_{PU}$ & 41.83$_{\pm0.24}$ & 48.83$_{\pm0.43}$ & 39.79$_{\pm0.54}$ & 74.32$_{\pm0.34}$ & 57.48$_{\pm0.36}$ & 54.63$_{\pm0.32}$ & 44.56$_{\pm0.20}$ & 81.71$_{\pm0.45}$ \\
\method & 42.21$_{\pm0.15}$ & 39.79$_{\pm0.37}$ & 27.68$_{\pm0.12}$ & 81.70$_{\pm0.19}$ & 63.00$_{\pm0.29}$ & 61.55$_{\pm0.41}$ & 67.56$_{\pm0.24}$ & 59.02$_{\pm0.30}$ \\ \midrule
\multicolumn{9}{c}{\textbf{Human + Distant}} \\ \midrule
DREEAM & 79.29$_{\pm0.23}$ & 78.89$_{\pm0.33}$ & 74.61$_{\pm0.24}$ & 85.15$_{\pm0.30}$ & 81.67$_{\pm0.35}$ & 78.95$_{\pm0.31}$ & 75.72$_{\pm0.35}$ & 84.88$_{\pm0.42}$ \\
ATLOP & 75.87$_{\pm0.53}$ & 74.83$_{\pm0.16}$ & 70.81$_{\pm0.73}$ & 80.67$_{\pm0.92}$ & 77.31$_{\pm0.65}$ & 75.72$_{\pm0.22}$ & 69.92$_{\pm0.26}$ & 84.10$_{\pm0.55}$ \\
KD-DocRE & 78.62$_{\pm0.56}$ & 77.15$_{\pm0.31}$ & 80.89$_{\pm0.24}$ & 73.02$_{\pm0.26}$ & 80.62$_{\pm0.45}$ & 80.32$_{\pm0.42}$ & 83.58$_{\pm0.21}$ & 75.06$_{\pm0.32}$ \\
SSR-PU & 80.09$_{\pm0.74}$ & 78.26$_{\pm0.30}$ & 74.51$_{\pm0.25}$ & 84.83$_{\pm0.30}$ & 80.52$_{\pm0.43}$ & 78.84$_{\pm0.31}$ & 74.24$_{\pm0.44}$ & 87.96$_{\pm0.51}$ \\
\method$_{PU}$ & 77.10$_{\pm0.61}$ & 76.85$_{\pm0.41}$ & 73.39$_{\pm0.15}$ & 80.94$_{\pm0.23}$ & 79.24$_{\pm0.34}$ & 80.99$_{\pm0.24}$ & 78.90$_{\pm0.50}$ & 80.49$_{\pm0.28}$ \\
\method & 83.56$_{\pm0.42}$ & 83.01$_{\pm0.35}$ & 88.09$_{\pm0.31}$ & 81.78$_{\pm0.27}$ & 84.01$_{\pm0.21}$ & 83.11$_{\pm0.37}$ & 86.03$_{\pm0.34}$ & 82.09$_{\pm0.27}$ \\ \bottomrule
\end{tabular}
}
\vspace{-1em}
\end{table*}
\subsection{Baselines}
For ReDocRED, we compared baselines ranging from fully supervised to distantly supervised. 
We will compare three settings, all evaluated on the same human-annotated test and development set. All models were chosen using the best scores on the dev set.

\begin{enumerate}[leftmargin=*]
\item \textbf{Human Annotated Only}: Denotes training only on the 3053 training dataset

\item \textbf{Distantly Supervised Only}: Denotes training only on the 101,873 distant dataset per DocRED, as ReDocRED does not revise this dataset.

\item \textbf{Human Annotated + Distantly Supervised}: Combines these two datasets. 
\end{enumerate}

Baseline models do this in a variety of different ways, with some using knowledge distillation (i.e. teacher training on human-annotated, student training from teacher output on distantly supervised). In \method, we fine-tune on distantly supervised training data via the regular loss, freeze the memory tokens, and then fine-tune on the training set.

For fully supervised (Human Annotation Only setting), we compare against ATLOP \cite{zhou2021document}, DREEAM \cite{ma2023dreeam}, KD-DocRE \cite{tan2022document}, and SSR-PU \cite{wang2022unified}. For the distantly supervised setting, we only compare against SSR-PU, as it is shown to be better than ATLOP. Furthermore, DREAM and KD-DocRE both primarily use knowledge distillation to achieve their improvements over ATLOP, only using the distantly supervised data to create the teacher logits to supervise the student model. Therefore, we believe that our "fine-tuning on both the distantly and fully supervised training data" would not maintain the spirit of the baseline method. Finally, since the main focus of many of the previous baselines combines distantly supervised work with human annotations, we also evaluate all models on the combined human+distantly supervised datasets.

For ChemDisGene, we compared against baselines BRAN \cite{verga2018simultaneously}, PubmedBert \cite{gu2021domain}, PubmedBert + BRAN \cite{zhang2022distant}, ATLOP, and \citet{wang2022unified} with (Positive-negative, positive-unlabelled, and the final SSR-PU variants).

In our experiments, we use precision, recall, and F1 scores as the evaluation metrics for the performance comparison. All standard deviations were calculated with 5 runs with different random seeds. More details about these evaluation metrics can be found in Appendix~\ref{app:Evaluation_Metrics}.

\section{Results}
\paragraph{Main Results}
Table~\ref{tab:results} shows the main results of our experiments. 
We see here that the Human Annotated Only dataset performs on the same level as the current SOTA results, DREEAM and SSR-PU.
However, we see that on the 2 other settings, Distantly Supervised and the combined Human + Distantly Supervised, \method outperforms other methods by a significant margin, when considering the standard deviation (+9 F1 and +3 F1 respectively). This implies that our model is much more effective with larger-scale training data even with noise in it, such as the distantly supervised training datasets. This intuitively makes sense since the memory tokens are initialized from scratch, and would benefit much more from larger-scale training data. Further research should seek to improve the initialization of the memory tokens, which could lead to faster training and further performance gains.

Finally, we observe that other baselines do not generally improve significantly even after training with distantly supervised and human-annotated data, which could be caused by architectural limits. Notably, this implies that \method's memory module adds processing capability that is actually \textit{significantly useful} (Section~\ref{sec:abl_deberta} shows us an example where adding more parameters does not help).

\begin{table}[t]
\centering
\caption{F1 on ChemDisGene test dataset (all relationships). The models shown with $^*$ are taken from \citet{wang2022unified} accordingly. Standard deviations are shown with 5 random seed runs. Note that all baseline results are from \citet{wang2022unified} and \citet{zhang2022distant}.}  
\label{tab:ablation_chem}
\resizebox{\linewidth}{!}{%
\begin{tabular}{cccc} 
Model      & F1         & Precision          & Recall          \\ \midrule
BRAN$^*$       & 32.5             & 41.8             & 26.6             \\
PubmedBert$^*$ & 42.1             & 64.3             & 31.3             \\
BRAN*       & 43.8             & 70.9             & 31.6             \\ \midrule
ATLOP$^*$      & 42.73$_{\pm 0.36}$ & \textbf{76.17$_{\pm 0.54}$} & 29.7$_{\pm 0.36}$  \\
PN         & 44.25$_{\pm 0.24}$ & 73.46$_{\pm 0.95}$ & 31.67$_{\pm 0.16}$ \\
PU         & 44.6$_{\pm 0.70}$  & 46.56$_{\pm 1.17}$ & 42.8$_{\pm 0.35}$  \\
SSR-PU     & 48.56$_{\pm 0.23}$ & 54.27$_{\pm 0.40}$  & 43.93$_{\pm 0.32}$ \\
\method     &\textbf{ 53.59$_{\pm 0.27}$} & 53.83$_{\pm 0.85}$ & \textbf{53.34$_{\pm 0.15}$} \\ \bottomrule
\end{tabular}
}
\vspace{-1em}
\end{table}

\paragraph{ChemDisGene Results} Table~\ref{tab:ablation_chem} shows that \method does indeed translate to other domains beyond the general task, with a 5 F1 point improvement over the best baseline. We observe \method performs well on the human-annotated training data. This is presumably because ChemDisGene has a larger dataset for training, so the memory tokens can be learned more effectively and, therefore it does not negatively affect performance as compared to the ReDocRED fully supervised setting.

\begin{table}[t]
\centering
\caption{Performance comparison of Top-K most common labels on the test dataset of ReDocRED. All but Top-K indicates the remainder of the $96-K$ labels.}
\label{tab:ablation_freq}
\resizebox{\linewidth}{!}{
\begin{tabular}{ccccc}
Model & F1 & Ign F1 & Precision & Recall \\ \toprule
\multicolumn{5}{c}{\textbf{Top 10 Labels}} \\ \midrule
ATLOP* & 62.12$_{\pm0.50}$ & 58.53$_{\pm0.62}$ & 50.70$_{\pm0.61}$ & 80.19$_{\pm0.45}$ \\
SSR-PU & 64.28$_{\pm0.31}$ & 60.87$_{\pm0.68}$ & 53.41$_{\pm0.35}$ & 80.72$_{\pm0.43}$ \\
SSR-PU+TTM & 68.21$_{\pm0.20}$ & 64.52$_{\pm0.43}$ & 57.94$_{\pm0.53}$ & 86.40$_{\pm0.34}$ \\ \midrule
\multicolumn{5}{c}{\textbf{All but Top 10 Labels}} \\ \midrule
ATLOP* & 39.47$_{\pm0.59}$ & 37.42$_{\pm0.34}$ & 27.34$_{\pm0.72}$ & 70.97$_{\pm0.60}$ \\
SSR-PU & 39.37$_{\pm0.41}$ & 37.39$_{\pm0.45}$ & 27.62$_{\pm0.62}$ & 68.51$_{\pm0.47}$ \\
SSR-PU+TTM & 44.04$_{\pm0.41}$ & 40.97$_{\pm0.49}$ & 32.01$_{\pm0.69}$ & 76.07$_{\pm0.69}$ \\ \midrule
\multicolumn{5}{c}{\textbf{Top 5 Labels}} \\ \midrule
ATLOP* & 55.79$_{\pm0.42}$ & 51.34$_{\pm0.29}$ & 42.41$_{\pm0.56}$ & 81.48$_{\pm0.32}$ \\
SSR-PU & 58.32$_{\pm0.32}$ & 54.01$_{\pm0.36}$ & 45.04$_{\pm0.79}$ & 82.69$_{\pm0.43}$ \\
SSR-PU+TTM & 62.03$_{\pm0.71}$ & 56.68$_{\pm0.28}$ & 48.78$_{\pm0.69}$ & 87.77$_{\pm0.35}$ \\ \midrule
\multicolumn{5}{c}{\textbf{All but Top 5 Labels}} \\ \midrule
ATLOP* & 47.60$_{\pm0.47}$ & 45.63$_{\pm0.60}$ & 35.63$_{\pm0.47}$ & 71.69$_{\pm0.45}$ \\
SSR-PU & 47.35$_{\pm0.54}$ & 45.46$_{\pm0.37}$ & 35.99$_{\pm0.54}$ & 69.20$_{\pm0.50}$ \\
SSR-PU+TTM & 53.02$_{\pm0.53}$ & 49.97$_{\pm0.31}$ & 41.17$_{\pm0.56}$ & 76.97$_{\pm0.40}$ \\ \bottomrule
\end{tabular}
}
\vspace{-1em}
\end{table}

\paragraph{Classifying Frequent/Infrequent Labels}
Previous work has shown that adding a memory component yields better performance on long-tailed or imbalanced class classification problems. We do generally see this phenomenon in Table~\ref{tab:ablation_freq}, as the difference is around 4 F1 and 4.5 F1 on the top 10 labels and the rest of the data. This difference is slightly more pronounced in the top 5, as we see a difference of 3.5 F1 and 5.5 F1 respectively. This indicates that baseline models perform slightly worse on the infrequent classes, whereas \method's memory component can help alleviate this performance drop.

\begin{table}[t]
\centering
\caption{Extremely unlabeled scenario (with less than 19\% of the original training labels as proposed by \citet{wang2022unified}). Standard deviations are shown with 5 random seed runs.}
\label{tab:ablation_ext}
\resizebox{\linewidth}{!}{%
\fontsize{15}{15}\selectfont
\begin{tabular}{ccccc}
Model      & F1               & Ign F1           & Precision        & Recall           \\ \midrule
\multicolumn{5}{c}{\textbf{Human Annotation Only}}                                   \\ \midrule
ATLOP     & 18.17$_{\pm0.66}$ & 18.14$_{\pm0.33}$ & 91.67$_{\pm3.39}$ & 11.16$_{\pm1.38}$ \\
SSR-PU     & 52.78$_{\pm0.46}$ & 51.53$_{\pm0.41}$ & 46.12$_{\pm0.57}$ & 61.69$_{\pm0.78}$ \\
\method & 52.60$_{\pm0.42}$ & 51.30$_{\pm0.43}$ & 43.97$_{\pm0.66}$ & 65.45$_{\pm0.58}$ \\ \midrule
\multicolumn{5}{c}{\textbf{Human Annotation + Distantly Supervised}}                                                    \\ \midrule
ATLOP     & 19.16$_{\pm0.23}$ & 19.01$_{\pm0.46}$ & 96.36$_{\pm0.41}$ & 8.51$_{\pm0.29}$  \\
SSR-PU     & 54.34$_{\pm0.21}$ & 54.12$_{\pm0.50}$ & 87.40$_{\pm0.61}$ & 39.43$_{\pm1.01}$ \\
\method & 66.47$_{\pm0.18}$ & 66.04$_{\pm0.46}$ & 81.40$_{\pm0.85}$ & 56.17$_{\pm0.94}$ \\ \bottomrule
\end{tabular}
}
\end{table}

\paragraph{Extremely Unlabeled Setting}
\citet{wang2022unified} introduced an "extremely unlabeled" scenario, that reduced the training labels to a mere 19\% of the original labels. 
We also evaluate our model on the extremely unlabeled setting (19\%) of the original training triples in ReDocRED \cite{wang2022unified} (Table~\ref{tab:ablation_ext}). We again see that \method does not work better than baselines on fully supervised, yet increases to 12 F1 points over the best baseline when allowed to train on distantly supervised data. We hypothesize that this is due to the better learning of the infrequent classes as shown in Table~\ref{tab:ablation_freq}.

\begin{figure}[t]
    \centering
    \includegraphics[width=\linewidth]{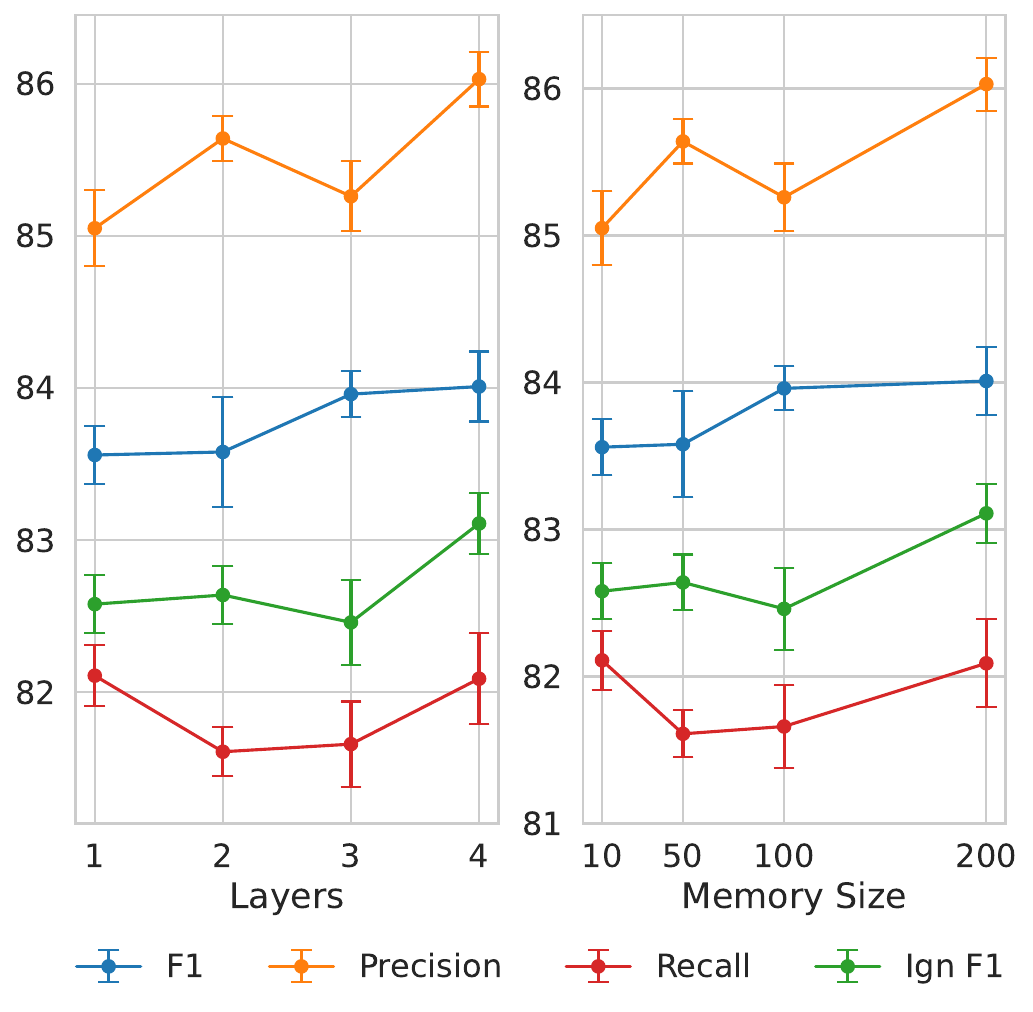}
    \caption{\textbf{Left Figure}: Effect of the size of the number of layers in the memory encoder. More layers imply a more powerful memory module. \textbf{Right Figure}: Effect of the number of memory tokens (Memory Size) available to be used in \method on the test dataset of ReDocRED. }
    \label{fig:ablation_layers_mem}
    \vspace{-1em}
\end{figure}

\paragraph{Memory Token Size} In Figure~\ref{fig:ablation_layers_mem}, we generally see an improvement in model performance (F1, Precision, and Ign F1) when increasing the READ module as well as the memory token size of the Token Turing Machine. While we halted at 4 layers and 200 tokens due to computational constraints, this trend is promising as it suggests that there are potential performance improvements awaiting exploration in future endeavors, albeit requiring increased computational resources.

\begin{table}[t]
\centering
\caption{Comparison of RoBERTa-large vs the more recent DeBERTaV3-large on the test dataset of ReDocRED.}
\label{tab:ablation_deberta}
\resizebox{\linewidth}{!}{%
\fontsize{15}{15}\selectfont
\begin{tabular}{ccccc}
Model  & F1               & Ign F1           & Precision        & Recall           \\ \midrule
\multicolumn{5}{c}{\textbf{Human Annotation Only (RoBERTa-large)}}                          \\ \midrule
SSR-PU & 80.18$_{\pm 0.31}$ & 78.61$_{\pm 0.46}$ & 69.43$_{\pm 0.43}$ & 90.50$_{\pm 0.53}$ \\
\method & 79.95$_{\pm 0.13}$ & 78.20$_{\pm 0.34}$ & 85.81$_{\pm 0.55}$ & 76.68$_{\pm 0.22}$ \\ \midrule
\multicolumn{5}{c}{\textbf{Human Annotation Only (DeBERTaV3-large)}}                        \\ \midrule
SSR-PU & 78.73$_{\pm 0.25}$ & 77.05$_{\pm 0.12}$ & 74.25$_{\pm 0.16}$ & 83.79$_{\pm 0.10}$ \\
\method & 78.88$_{\pm 0.22}$ & 77.29$_{\pm 0.19}$ & 75.63$_{\pm 0.21}$ & 82.42$_{\pm 0.18}$ \\ \midrule
\multicolumn{5}{c}{\textbf{Human + Distant (RoBERTa-large)}}                                \\ \midrule
SSR-PU & 80.52$_{\pm 0.43}$ & 78.84$_{\pm 0.31}$ & 74.24$_{\pm 0.44}$ & 87.96$_{\pm 0.51}$ \\
\method & 84.01$_{\pm 0.21}$ & 83.11$_{\pm 0.37}$ & 86.03$_{\pm 0.34}$ & 82.09$_{\pm 0.27}$ \\ \midrule
\multicolumn{5}{c}{\textbf{Human + Distant (DeBERTaV3-large)}}                              \\ \midrule
SSR-PU & 79.65$_{\pm 0.27}$ & 78.34$_{\pm 0.23}$ & 82.65$_{\pm 0.27}$ & 78.49$_{\pm 0.19}$ \\
\method & 80.56$_{\pm 0.16}$ & 79.49$_{\pm 0.21}$ & 83.44$_{\pm 0.24 }$& 77.88$_{\pm 0.25}$ \\ \bottomrule
\end{tabular}
}
\vspace{-1em}
\end{table}

\paragraph{Using DebertaV3 as the Base Model:} 
\label{sec:abl_deberta} Interestingly, all baselines generally rely on Roberta-large as the base model. We also explored using DebertaV3-large, which is presented as a more recent and powerful model due to its larger parameter count and higher performance on the GLUE benchmark (improvements include disentangling attention, an enhanced decoding layer \cite{he2020deberta}, and Electra-style pretraining \cite{he2021debertav3}). However, from Table~\ref{tab:ablation_deberta}, we see that for document RE, it surprisingly \textit{does not improve} performance. Because of this observation, the \method also uses Roberta-large. 
Additionally, this demonstrates a case where naively adding parameters \textit{does not} help improve relation classification performance, whereas adding the memory mechanism \textit{does}.

\section{Discussion}
\textbf{Analysis of Memory Module} The Token Turing Machine (TTM) memory module performs best with a large training set. This leads to an important question: what other applications could provide such extensive training data? While it is true that TTM benefits from a large dataset, potential applications include any field within NLP or computer vision. Moreover, these datasets do not necessarily need to be human-annotated. For instance, the DocRED dataset was distantly supervised using Wikipedia and spaCy for NER and relation linking via Wikidata. Similar methods can be employed to create datasets for specific tasks. TTM-RE has demonstrated improved performance with these distantly supervised datasets compared to other baselines.

\textbf{Dataset Size} We hypothesize that performance is correlated with the size of the dataset. Notably, TTM-RE outperforms baselines in ChemDisGene without finetuning on a related dataset. Given that the human-annotated ReDocRED dataset contains only 3,053 documents, compared to 101,873 for distantly supervised datasets and 76,942 for ChemDisGene, the memory mechanism may require a certain amount of finetuning before it is fully effective. This suggests further research is needed to find more efficient ways to optimize the memory mechanism. However, future work is required to fully investigate this phenomenon.

\section{Conclusion}

In this paper, we investigated \method, integrating TTMs relation classification and evaluated our model on ReDocRED and ChemDisGene RE datasets.
To summarize our contributions, no previous work has explored memory in this distantly supervised setting. As such, \method demonstrates a completely new way of increasing performance for the difficult task of document-level relation extraction as opposed to previous work, which mainly improved the loss function \cite{zhou2021documentadaptiveThre, wang2022unified}. For this work, we found compelling results by performing ablations that showed that adding (and increasing) the number of memory tokens/layers helped performance over baselines, compared to simply using larger models like Deberta V3. Table~\ref{tab:ablation_freq} also demonstrates the improvement in the less-represented labels as opposed to the top 10 labels (which comprise 62\% of the dataset). Additionally, we show that in Figure~\ref{fig:ablation_layers_mem}, performance continues to improve as we add more memory tokens.

We also observed that TTMs necessitate either fine-tuning on a large distantly-labeled training dataset or a significantly large human-annotated training dataset (ChemDisGene) to optimize memory vector initialization. 
We believe that this work lends itself to future work in this exciting area, and we hope that our findings will pave the way for future exploration of memory-augmented techniques in large language models for information extraction tasks. 

\section*{Limitations}
Although we investigated multiple different LLMs and parameters and the type relation distribution for relation prediction as well as addressing the false positives, the performance we attained is still limited compared to supervised methods on the same task. 
Relation prediction still requires a large amount of data, despite \method's ability to use distantly supervised data. Future work should seek to tackle this approach that combines labeled data creation with SOTA document relation extraction models for maximum efficiency on human annotators.

\section*{Ethical Statement}
Based on the methodology we have currently employed, we do not foresee any significant ethical concerns. 
All the documents and models utilized in our study were obtained from open-source domains, ensuring a transparent and accessible source of information. 
Additionally, \method is trained on purely open-source document relation extraction data, eliminating the risk of privacy leakage.
Additionally, the task of relation extraction is a widely recognized and well-studied problem across various natural language processing applications.

However, it is crucial to acknowledge a minor factor, namely the presence of potential hidden biases within the pre-trained language models used in our analysis. 
These biases may stem from the data on which the models were trained, which could have inadvertently introduced implicit human biases. 
While our usage of these pre-trained language models enables us to identify relationships between arbitrary entities, it is conceivable that biases may emerge if one were to explore sensitive relation classes and entities.

ChatGPT and Grammarly were used for parts of the writing.
In total, training took more than 75 hrs on NVIDIA RTX A6000 for pretraining in total. The main roadblock was the distantly supervised finetuning portion for all of the models, due to the size of the dataset.
Derivatives of data accessed for research purposes should not be used outside of research contexts. Code will be released at \url{https://github.com/chufangao/TTM-RE}.

\bibliography{custom}
\bibliographystyle{acl_natbib}

\clearpage
\appendix

\section{Parameter Settings}
\label{app:hyperparameters}
All models were run on an NVIDIA A6000 with 48 gigabytes of VRAM. Still, around 10 days were required to fully run the experiments. For particularly expensive computations, like $Logits_{SR}$, only the fastest model--UnifiedQA-large--could be feasibly run.

All models were downloaded from Huggingface \cite{wolf2019huggingface}. We used the default setup of the pre-trained models and did not do further finetuning. All the step mentioned in the methodology section works on the output of the pre-trained models.

Supervised results DREEEAM \cite{ma2023dreeam} and KD-DocRE \cite{tan2022documentAdaptiveFocalLoss} were taken from the original source papers. 

\section{Evaluation Metrics}
\label{app:Evaluation_Metrics}
To keep in tradition with existing document relation extraction work, we report both F1 and Ign\_F1 as computed by the official metrics from ReDocRED. 
F1 refers to micro-averaged F1 score that combines precision $P$ and recall $R$ $$F1 = \frac{2 P R}{P+R}$$
$$P = \frac{\texttt{length of correct (h,t,rel) preds}}{\texttt{length of all  (h,t,rel) preds}}$$
$$R = \frac{\texttt{length of correct (h,t,rel) preds}}{\texttt{length of correct  (h,t,rel)}}$$
Where \texttt{(h,t,rel)} denotes a tuple of the predicted head, tail, and relation.
Ign\_F1 is computed similarly to above but ignores the samples in the DocRED's distantly supervised training set. (Note that we do not use any distantly labeled data).

\section{Ablation Tables}
Tables~\ref{tab:ablation_layers} and \ref{tab:ablation_mem} for Figure~\ref{fig:ablation_layers_mem}.

\begin{table}[ht]
\centering
\caption{Ablation regarding the number of layers in the memory encoder. More layers imply a more powerful memory module. Results are evaluated from the test dataset of ReDocRED.}
\label{tab:ablation_layers}
\resizebox{\linewidth}{!}{%
\fontsize{15}{15}\selectfont
\begin{tabular}{ccccc}
Num Layers & F1               & Ign F1           & Precision         & Recall           \\ \midrule
1                & 83.56$_{\pm 0.19}$ & 82.58$_{\pm 0.19}$ & 85.05$_{\pm 0.25}$ & 82.11$_{\pm 0.20}$ \\
2                & 83.58$_{\pm 0.36}$ & 82.64$_{\pm 0.19}$ & 85.64$_{\pm 0.15}$  & 81.61$_{\pm 0.16}$ \\
3                & 83.96$_{\pm 0.15}$ & 82.46$_{\pm 0.28}$ & 85.26$_{\pm 0.23}$  & 81.66$_{\pm 0.28}$ \\
4                & 84.01$_{\pm 0.23}$ & 83.11$_{\pm 0.20}$ & 86.03$_{\pm 0.18}$  & 82.09$_{\pm 0.30}$ \\ \bottomrule
\end{tabular}
}
\end{table}

\begin{table}[ht]
\centering
\caption{Effect of the size of the number of memory tokens available to be used in \method on the test dataset of ReDocRED. }
\label{tab:ablation_mem}
\resizebox{\linewidth}{!}{%
\fontsize{15}{15}\selectfont
\begin{tabular}{ccccc}
Mem. Size & F1               & Ign F1           & Precision        & Recall           \\ \midrule
10                & 83.20$_{\pm 0.23}$ & 82.24$_{\pm 0.16}$ & 85.31$_{\pm 0.12}$ & 81.18$_{\pm 0.13}$ \\
50                & 83.52$_{\pm 0.21}$ & 82.59$_{\pm 0.14}$ & 85.65$_{\pm 0.31}$ & 81.50$_{\pm 0.17}$ \\
100                & 83.67$_{\pm 0.19}$ & 82.82$_{\pm 0.22}$ & 87.01$_{\pm 0.24}$ & 80.57$_{\pm 0.20}$ \\
200                & 84.01$_{\pm 0.17}$ & 83.11$_{\pm 0.20}$ & 86.03$_{\pm 0.19}$ & 82.09$_{\pm 0.13}$ \\ \bottomrule
\end{tabular}
}
\end{table}
\section{PCA Plots of Memory Tokens} 
\begin{figure}[ht]
    \centering
    \includegraphics[width=\linewidth]{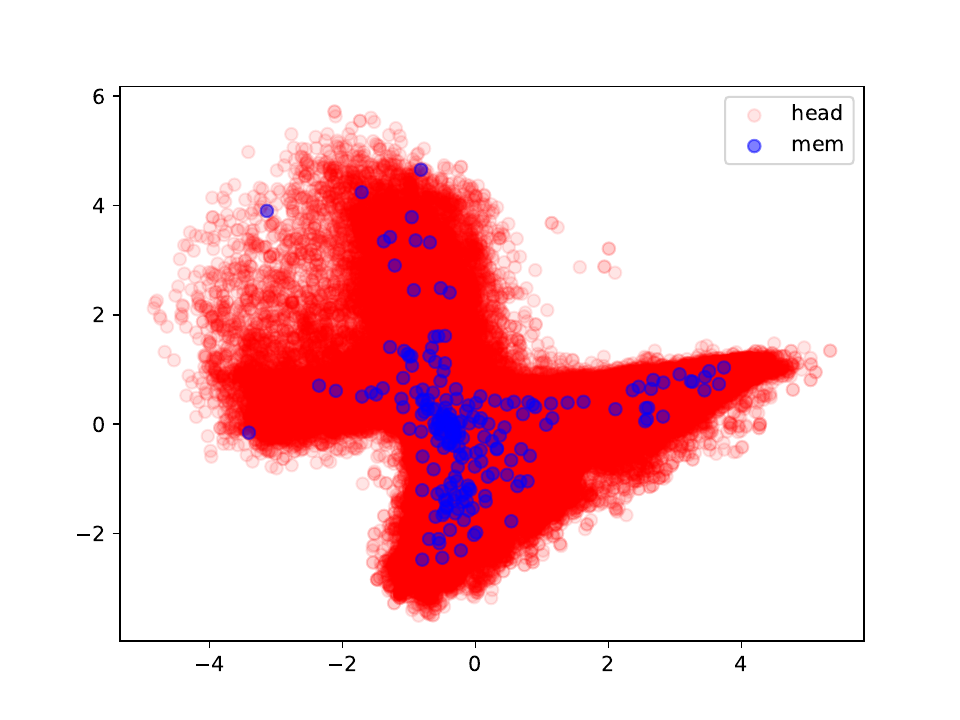}
    \caption{Plot of PCA-transformed head entities along with (200) memory entities. Tail entities are omitted due to redundancy.}
    \label{fig:mem_pca}
\end{figure}

Shown in Figure~\ref{fig:mem_pca}, we see that the token embeddings lie in a scattered space around the head entities. This makes intuitive sense as the prototypical tokens should capture a diverse set of different types of head tokens.

\section{Literature Review Continued}

Pre-trained language models, such as BERT-based architectures \cite{xu2021entityPLMBased}, have shown considerable efficacy in document-level relation extraction. BERT-based methodologies have integrated approaches like hierarchical inference networks \cite{tang2020hin}, enhanced co-reference reasoning \cite{ye2020coreferential}, and adaptive thresholding. Furthermore, graphical neural networks (GNNs) \cite{zeng2020doublegraph} have been leveraged for document-level relation extraction, employing techniques such as feature learning on a coreference graph \cite{sahu2019inter}, edge-oriented learning strategies \cite{christopoulou2019connecting}, attention mechanisms \cite{guo2019attention}, and iterative refinement methods to aggregate multi-hop information \cite{nan2020reasoning}.

\subsection{Weakly Supervised Document Relation Extraction}
Past studies in document-level relation extraction have heavily depended on human annotation to create training datasets, a process known for its time-consuming and labor-intensive nature. There has been minimal exploration into document relation extraction methods that do not necessitate human annotation.

Various weakly supervised methods have been extensively investigated for relation extraction \cite{jiang2009multi, huang2017deep, qu2018weakly, wang2018open, li2018pattern}. For instance, \citet{huang2017deep} employed residual connections and convolutional neural networks (CNNs) to identify pertinent candidates, thereby enhancing supervised relation classification. In a similar vein, \citet{qu2018weakly} extracted textual patterns from initial examples to offer supplementary supervision. Introducing a ranking-based approach for seed selection, \citet{phi2018ranking} improved bootstrapping and distantly supervised relation extraction. Additionally, \citet{sainz2021label} proposed representing each relation class using a label verbalizer and tackled the relation extraction task with a textual entailment model. 

Moreover, \citet{wang2022unified} showed an "extremely unlabeled" scenario wherein each relation type comprised only one instance, consequently reducing the training set to a smaller number of labeled relation triplets. However, this scenario does not help in \textit{improving} performance on the fully supervised test data overall.

\citet{qu2018weakly} derived textual patterns from initial samples and employed them as weak signals for relation extraction. 
\citet{gao2023promptre} investigated purely weakly-supervised prompting methods devoid of human labels and revealed significant limitations in relying solely on weak supervision, particularly in handling a high incidence of hallucinations when predicting no-relation entity-entity pairs. 







\section{SSR-PU Loss}
\label{app:ssrpu_loss}
\subsection{Class-shift Adjusted Positive Unlabeled Loss Function (SSR-PU)}
Previous supervised document-level RE methods only treated unlabeled relations as negative samples, which may lead to low recall in the presence of a large number of false negatives. To address this problem, we adopt PU learning with prior shift similar to \citet{wang2022unified} for each class \cite{pmlr-v37-plessis15, NIPS2014_35051070}.

PU learning assumes that unlabeled data can reflect the true overall distribution, that is, $p_{\mathrm{U}_{i}}(\boldsymbol{x})=p_{i}(\boldsymbol{x})$. The expected classification risk formulation can be defined in a form that can be approximated using the data like so:
\begin{equation}
\begin{aligned}\label{eq7}
\widehat{R}_{\mathrm{PU}}(f) &=\sum_{i=1}^{K}(\frac{\pi_{i}}{n_{\mathrm{P}_{i}}} \sum_{j=1}^{n_{\mathrm{P}_{i}}} \ell(f_{i}( \boldsymbol{x}_{j}^{\mathrm{P}_{i}}), +1) \\
&+\mathrm{max}(0,[\frac{1}{n_{\mathrm{U}_{i}}} \sum_{j=1}^{n_{\mathrm{U}_{i}}} \ell(f_{i}( \boldsymbol{x}_{j}^{\mathrm{U}_{i}}), -1)\\
&-\frac{\pi_{i}}{n_{\mathrm{P}_{i}}} \sum_{j=1}^{n_{\mathrm{P}_{i}}} \ell(f_{i}( \boldsymbol{x}_{j}^{\mathrm{P}_{i}}), -1)]))
\end{aligned}
\end{equation}
where $\pi_{i} = p(y_i = +1)$ denotes probability of positive prior for relation class $i$. $n_{P_i}$ are the number of positive and $n_{U_i}$ are the unlabelled samples of
class i, respectively. $\ell$ is a convex loss function, and $f_i(\cdot)$ is a score function that predicts class $i$. $\boldsymbol{x}_{j}^{\mathrm{P}_{i}}$ and $\boldsymbol{x}_{j}^{\mathrm{U}_{i}}$ denotes that the $j$-th sample of class $i$ is positive and unlabeled as class $i$ respectively. 
Note that without the max function, the second term in Eq.\ref{eq7} can be negative and can be prone to overfitting (and therefore highly negative) when using a highly flexible model. Thus, a non-negative risk component \citep{NIPS2017_7cce53cf} is used to solve the overfitting problem. Note that $n_{U_i}$ is essentially a hyperparameter that one assumes before model training ($n_{P_i}, \pi_i$ can be directly calculated by counting, and everything else is learned via backprop).

While the original method additionally corrected for the heavy class imbalance problem via multiplying $\gamma_{i}=\frac{1-\pi_{i}}{\pi_{i}})^{0.5}$ before positive risk estimations as the class weight, we found that this was unnecessary and that tuning the other hyper-parameters was sufficient in reproducing the original paper results.

\paragraph{Prior Shift:} Ordinary PU learning requires an assumption that the overall distribution needs to be the same as the distribution of the unlabeled data. In contrast, with the document-level RE dataset constructed by a \textit{recommend-revise} scheme, where \citet{wang2022unified} found that there existed a prior shift in the unlabeled data of the training data vs the test data. When these priors are different, ordinary PU learning will yield a biased result. 

To address this problem, inspired by the method \citep{charoenphakdee2019positive} for handling a prior shift between the test set and the training set, a correction term is added.
For each class, assume that the original prior $\pi_{i}=p(y_{i}=+1)$. Let $\pi_{labeled,i}=p(s_{i}=+1)$ and $(1-\pi_{labeled,i})=(1-p(s_{i}=+1))=p(s_{i}=-1)$ where $s_{i}=+1$ or $s_{i}=-1$ mean that the $i$-th class is labeled or unlabeled, respectively. 

The conditional probability of a positive sample under unlabeled data is:
\begin{equation}
\begin{aligned}
\pi_{u,i} &= p(y_{i}=1 \mid s_{i}=-1) \\
&= \frac{p(y_{i}=1, s_{i}=-1)}{p(s_{i}=-1)}\\
&= \frac{p(y_{i}=1)-p(s_{i}=+1)}{p(s_{i}=-1)}\\
&= \frac{\pi_{i}-\pi_{labeled,i}}{1-\pi_{labeled,i}}
\end{aligned}
\end{equation}
where Step 3 is true because positive samples are assumed to be a superset of the unlabelled data and negative data has no overlaps with the labeled data. I.e. $p(y_i=-1, s_i)=0$. 
Finally, the non-negative risk estimator \citep{NIPS2017_7cce53cf} under class prior shift of training data is obtained as follows:
\begin{equation}
\begin{aligned}
&\widehat{R}_{\mathrm{S-PU}}(f)=\sum_{i=1}^{K}( \frac{\pi_{i}}{n_{\mathrm{P}_{i}}} \sum_{j=1}^{n_{\mathrm{P}_{i}}}\ell(f_{i}(\boldsymbol{x}_{j}^{\mathrm{P}_{i}}), +1) \\&+\mathrm{max}(0, [\frac{1}{n_{\mathrm{U}_{i}}} \frac{1-\pi_{i}}{1-\pi_{u,i}} \sum_{j=1}^{n_{\mathrm{U}_{i}}} \ell(f_{i}(\boldsymbol{x}_{j}^{\mathrm{U}_{i}}), -1)\\&-\frac{1}{n_{\mathrm{P}_{i}}}\frac{\pi_{u,i}-\pi_{u,i} \pi_{i}}{1-\pi_{u,i}} \sum_{j=1}^{n_{\mathrm{P}_{i}}}\ell(f_{i}(\boldsymbol{x}_{j}^{\mathrm{P}_{i}}), -1)]))
\end{aligned}
\end{equation}

The proof is shown in Theorem 1 in \citet{wang2022unified}.

\section{Distribution of Labels}
\label{app:rel_dist}
\begin{figure*}
    \centering
    \includegraphics[width=\linewidth]{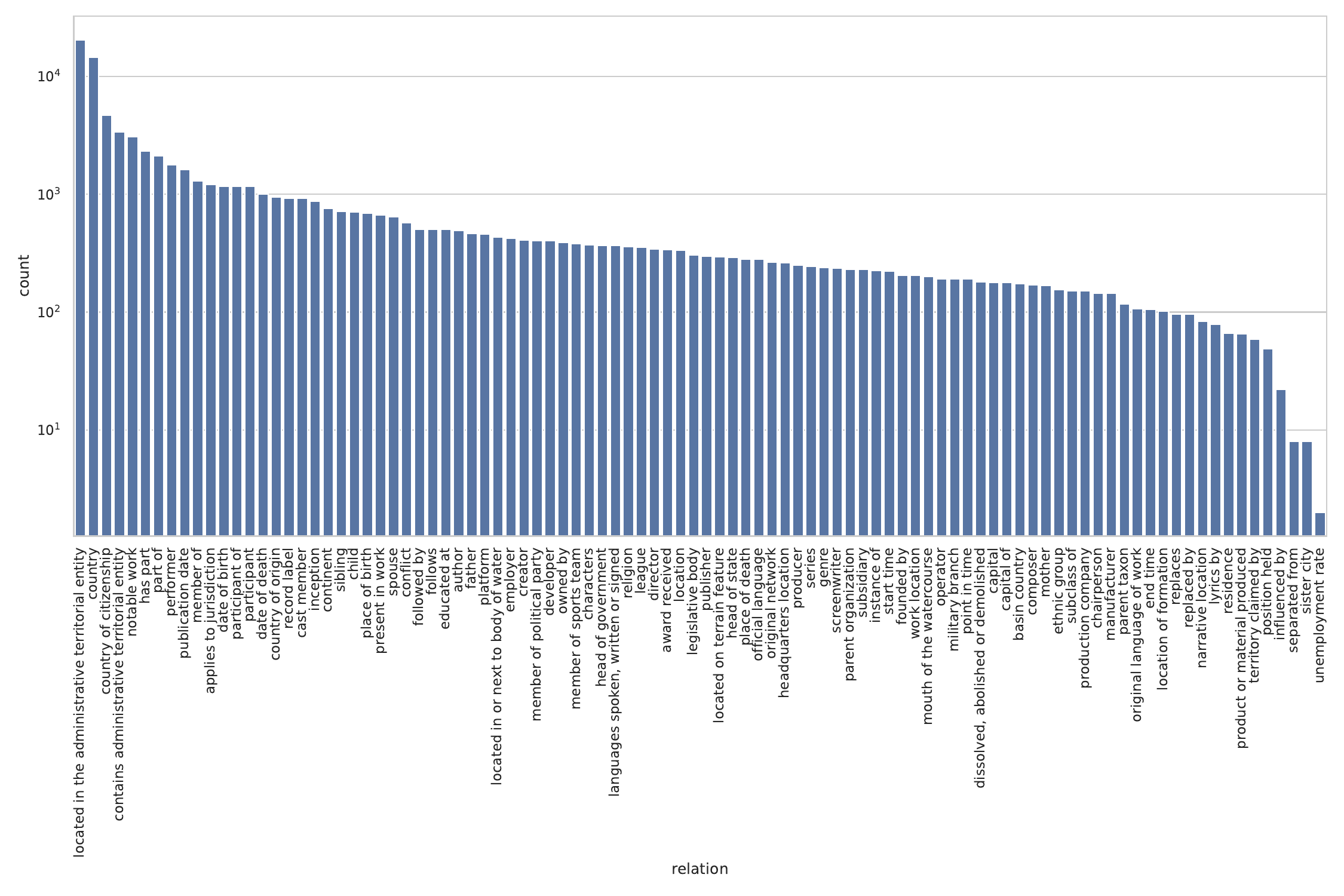}
    \caption{Distribution of All Relations in the Training set of ReDocRED.}
    \label{fig:rel_dist}
\end{figure*}

We visualize a distribution of labels to see which relations have the least amount of occurrences in Figure~\ref{fig:rel_dist}. We see that "unemployment rate", "sister city", and "separated from" are the smallest by a few orders of magnitude. This uneven label distribution makes document relation classification even harder.

\section{Case Study On Rare Events}
From Appendix~\ref{app:rel_dist}, we see that many labels are quite rare. What if we restrict our labels to only those with less than the 25th quartile? Then, we obtain a dataset that is 0.027 of the total relation labels in the test set. However, whenever \method predicts one of these relations, it has an 80\% chance to get it right. 

E.g. We were able to predict that "\texttt{Republic of China} is "\texttt{territory claimed by}" "\texttt{People's Republic of China}" (a relation in the 6th percentile) from this paragraph: 

\texttt{The " March of the Volunteers " is the national anthem of the People 's Republic of China , including its special administrative regions of Hong Kong and Macau .
Unlike most previous Chinese state anthems , it is written entirely in the vernacular , rather than in Classical Chinese .
Its lyrics were composed as a dramatic poem by the poet and playwright , the Japan - educated Tian Han in 1934 and set to music by Nie Er from Yunnan Province the next year for the film Children of Troubled Times .
It was adopted as the PRC 's provisional anthem in 1949 in place of the " Three Principles of the People " of the Republic of China and the Communist " Internationale " .
When Tian Han was imprisoned during the Cultural Revolution in the 1960s , the march was briefly and unofficially replaced by " The East Is Red " , then played without words , then played with altered words .
Restored to its original version , the " March of the Volunteers " was raised to official status in 1982 , adopted by Hong Kong and Macau upon their restorations to China in 1997 and 1999 , respectively , and included in the Chinese Constitution 's Article 136 in 2004 .
}

However, we note that this fact is not explicitly mentioned in this. It is possible that the memory mechanism enabled this prediction. Although we hypothesize that adding the memory gives it a larger context to be able to compare incoming entities, further research is needed to fully investigate the rare relation performance.
\end{document}